\documentclass[final]{cvpr}

\usepackage{times}
\usepackage{epsfig}
\usepackage{graphicx}
\usepackage{multirow}
\usepackage{amsmath}
\usepackage{amssymb}

\usepackage[pagebackref=true,breaklinks=true,colorlinks,bookmarks=false]{hyperref}

\def\cvprPaperID{11476} 
\def\confYear{CVPR 2022}

\begin{document}

\title{
BAPose: Bottom-Up Pose Estimation with Disentangled Waterfall Representations
}



\author{Bruno Artacho \hspace{2cm} Andreas Savakis\\
Rochester Institute of Technology\\
Rochester, NY\\
{\tt\small bmartacho@mail.rit.edu \hspace{0.5cm} andreas.savakis@rit.edu}}

\maketitle

\begin{abstract}
We propose BAPose, a novel bottom-up approach that achieves state-of-the-art results for  multi-person pose estimation. Our end-to-end trainable framework leverages a disentangled multi-scale waterfall architecture and incorporates 
adaptive convolutions to  infer keypoints more precisely in crowded scenes with occlusions.
The multi-scale representations, obtained by the disentangled waterfall module in BAPose, leverage the efficiency of progressive filtering in the cascade architecture, while maintaining multi-scale fields-of-view comparable to spatial pyramid configurations.
Our results on the challenging COCO and CrowdPose datasets demonstrate that BAPose is an efficient and robust framework for multi-person pose estimation, achieving significant improvements on state-of-the-art accuracy. 
\end{abstract}

\section{Introduction}
Locating humans and estimating their pose in crowded scenes is a challenging task of high interest for computer vision researchers and practitioners. Successful human pose estimation enables applications in action recognition, sports analysis, human-computer interactions, rehabilitation, and sign language recognition. Various methods have focused on tackling specific aspects of human pose estimation, including 2D pose estimation \cite{DeepPose}, \cite{HourGlass},
\cite{CPM}, \cite{UniPose},
\cite{HRNet}; 3D pose estimation \cite{LCR-Net}, \cite{Monocap}, \cite{DensePose}, \cite{UniPose+}; single frame detection \cite{RecurrentPose}; pose detection in videos \cite{3D_Video_Occlusion-Aware}; dealing with a single person \cite{CPM} or multiple people \cite{OpenPose}.


The task of multi-person pose estimation is notorious for the challenges caused by the high occurrence of joint occlusions, 
combined with the large number of degrees of freedom in the human body movements. Common approaches to overcome these challenges include the deployment of statistical and geometric models to estimate occluded joints \cite{GeometricPose}, \cite{StatisticalPose} 
and the use of anchor poses \cite{LCR-Net}, \cite{Point-setAnchors}, although the latter approach is limited by the number of poses in its library, making it difficult to generalize and handle unforeseen poses.


\begin{figure}[t]
\begin{center}
\includegraphics[width=1\linewidth]{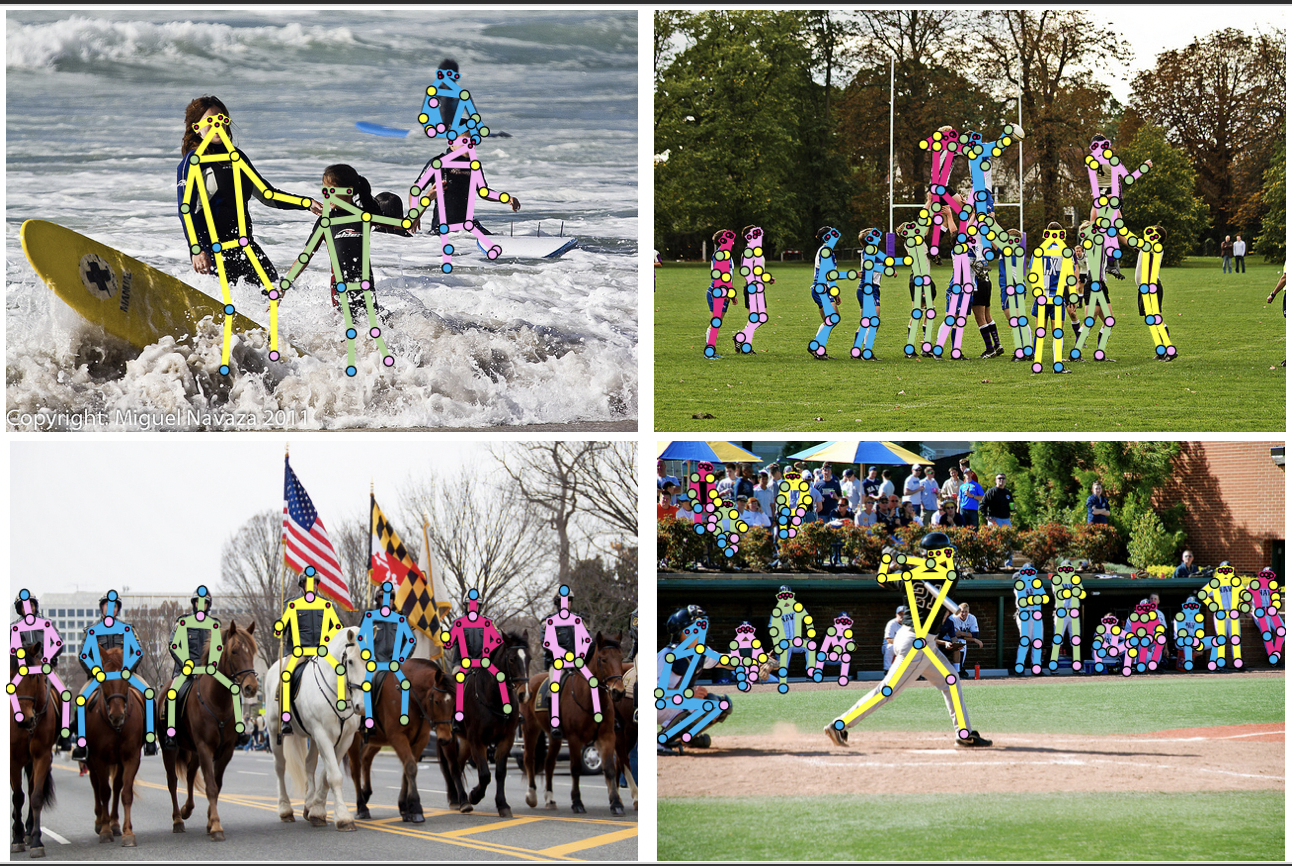}
\end{center}
  \caption{Pose estimation examples with our BAPose method.}
\label{fig:samples}
\end{figure}

State-of-the-art (SOTA) methods for multi-person 2D pose estimation can be divided in two distinct approaches: top-down and bottom-up. Top-down methods initially detect the instances of persons in the image and then perform single person pose estimation for each individual. Bottom-up methods for multi-person pose estimation either detect all keypoints and group them by affinity relations \cite{OpenPose}, \cite{Pifpaf}, or directly regress the keypoint locations to each person in the image \cite{DEKR}.
Overall, top-down approaches achieve high accuracy, although they require an extra step for person detection, resulting in a slower and more costly process. On the other hand, bottom-up approaches are based on a 
single-stage for multi-person pose estimation that is generally more efficient.

In this paper, we propose BAPose, a bottom up framework that is named after ``Basso verso l'Alto'' (bottom up in Italian).
The BAPose method is a single-stage, end-to-end trainable network
that extends recent successful approaches by UniPose \cite{UniPose}
, UniPose+ \cite{UniPose+}, 
and OmniPose \cite{OmniPose}
to bottom-up multi-person 2D pose estimation.
BAPose achieves state-of-the-art (SOTA) results  in two large datasets without requiring post-processing, intermediate supervision, multiple iterations or anchor poses.
The main contributions of BAPose are the following:
\begin{itemize}
\vspace{-0.1in}
\item We propose the novel BAPose method, a single-pass, end-to-end trainable, multi-scale approach for bottom-up multi-person 2D pose estimation, that achieves SOTA results for two large benchmark datasets, COCO and CrowdPose.
\vspace{-0.1in}
\item In our bottom-up approach, we combine multi-scale waterfall features 
with disentangled adaptive convolutions and an integrated multi-scale decoder to disambiguate the joints of individuals in crowded scenes.
\vspace{-0.1in}
\item The enhanced multi-scale capability of BAPose specializes the network for human pose estimation in images with a large number of person instances, drastically increasing the SOTA performance for the CrowdPose dataset.
\end{itemize}

\section{Related Work}\label{sec:related_work}
The advent of Convolutional Neural Networks (CNNs) for deep learning methods enabled leaping advances for the task of human pose estimation \cite{DeepPose}, \cite{DEKR}, \cite{OpenPose}, \cite{OmniPose}, \cite{LCR-Net}, \cite{UniPose+}.
The Convolutional Pose Machines (CPM) \cite{CPM} approach uses a sequence of CNN stages in the network to refine joint detection.
Furthering the work of \cite{CPM} Yan et al. integrated Part Affinity Fields (PAF) in their framework to better capture the context and relationships between joints for improved 2D human pose estimation. The resulting OpenPose method \cite{OpenPose} 
is widely used in various applications.

The Stacked Hourglass (HG) network \cite{HourGlass} utilizes a multi-stage approach by cascading hourglass structures through the network to refine the resulting pose estimation. The HG work was further expanded to incorporate the multi-context approach in 
\cite{Multi-context} by augmenting the backbone with residual units in order to increase the receptive Field-of-View (FOV). Postprocessing with Conditional Random Fields (CRFs) is applied to refine the location of detected joints. A downside of this approach is the increase in complexity by the addition of another stage of postprocessing and the associated increase in computational load.

Aiming to offer a multi-scale approach to feature representations, the High-Resolution Network (HRNet) includes both the high and low resolutions to obtain a larger FOV.
The Multi-Stage Pose Network (MSPN) \cite{MSPN} follows a similar approach to HRNet by combining the cross-stage feature aggregation and coarse-to-fine supervision.
In further work, \cite{HigherHRNet} combined the HRNet structure with multi-resolution pyramids to obtain multi-scale features.
Building upon the work of HRNet, the Distribution-Aware coordinate Representation of Keypoints (DARK) method \cite{DarkPose} incorporates a refined approach to their decoder in order to reduce the inference error at the decoder stage.

Developments in graphical components for CNNs inspired the Cascade Prediction Fusion (CPF) approach \cite{zhang2019human} that applies graphs in other to further extract the contextual information for pose estimation.
In a similar fashion, Cascade Feature Aggregation (CFA) \cite{su2019improvement} applied the cascade approach into the semantic information for pose estimation. Generative Adversarial Networks (GANs) were used in \cite{SAGAN} in order to learn dependencies and contextual information for pose.

A limitation of top-down approaches is the requirement of an independent module for the detection of instances of humans in the frame. LightTrack \cite{LightTrack}, for instance, requires a separate YOLOv3 \cite{YOLOv3} architecture to detect subjects prior to the detection of joints for pose estimation.
In a slightly different approach, LCR-Net \cite{LCR-Net} applies multiple branches for detection by using Detectron \cite{Detectron} and the arrangement of joints during classification.

With the goal of developing a unified framework to overcome the limitation of top-down approaches, UniPose \cite{UniPose} combines the bounding box generation and pose estimation in a single, one-pass network.
This approach is possible due to the larger FOV and significant increase in the multi-scale representation obtained by the Waterfall Atrous Spatial Pooling (WASP) module \cite{WASPnet}, which allows for greater FOV and results in better representation of contextual information.

\begin{figure*}[t]
\begin{center}
\includegraphics[width=1\linewidth]{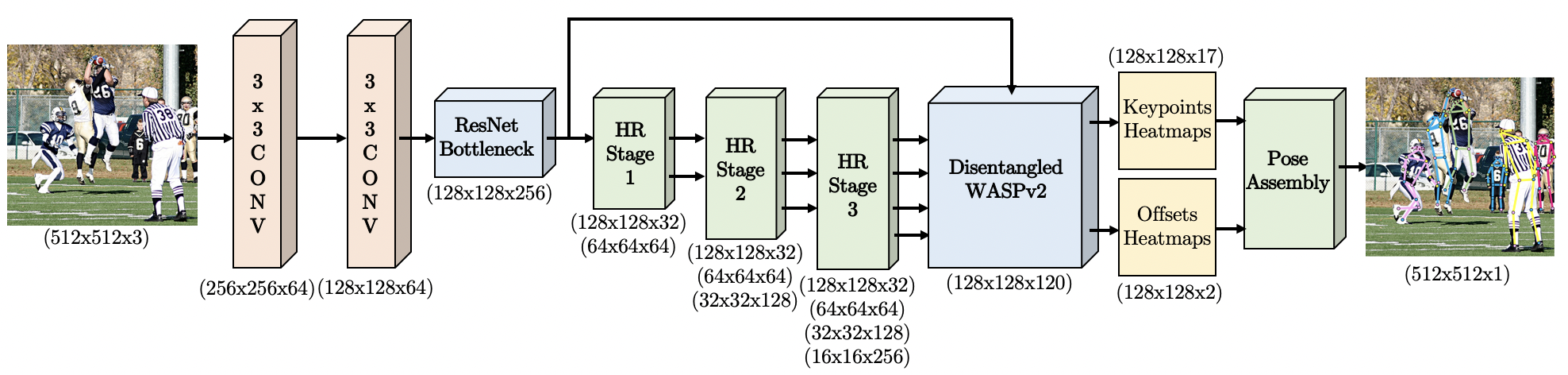}
\end{center}
  \caption{BAPose architecture for bottom-up multi-person pose estimation. The input color image is fed through the HRNet backbone for initial feature extraction, followed by the D-WASP module and an adaptive convolution based decoder to generate one heatmap per joint (17 joints in the figure) and offset regression for the localization of each person instance.}
\label{fig:BAPose}
\end{figure*}

\subsection{Bottom-Up Approaches}
The most common approach for bottom-up estimation is to associate detected keypoints with each person present in the image. One approach is to cast the problem in terms of integer linear programming \cite{DeepCut}, \cite{DeeperCut}. A clear drawback of this approach is the high processing time required, inhibiting the possibility of real-time performance.

OpenPose \cite{OpenPose} is considered a breakthrough approach for grouping keypoints with the introduction of PAF. Other methods further developed PAF, such as Pif-Paf \cite{Pifpaf} and  associative embedding \cite{AE}. 
In a similar vein,
PersonLab \cite{PersonLab} adopted Hough voting, and \cite{HGG} used hierarchical graphical clustering.

The dense regression of pose candidates is adopted by several recent works \cite{SSPM}, \cite{PPN}. A limitation of this approach is the lower regression accuracy in the localization process, that usually requires an additional post-processing step in order to improve the regression results. Aiming to bridge the gap, \cite{MDR} applied a mixture density network to better handle uncertainty in the network before regression. 
The recent Disentangled Keypoint Regression (DEKR) method \cite{DEKR}, on the other hand, learns disentangled representations for each keypoint and utilizes adaptively activated pixels, ensuring that each representation focuses on the corresponding keypoint area.

\subsection{Multi-Scale Feature Representations}
The reduction of resolution that takes place in CNN-based methods is an ongoing challenge for pose estimation or semantic segmentation methods.
Fully Convolutional Networks (FCN) \cite{FCN} initially addressed resolution reduction by adopting upsampling strategies across deconvolution layers to increase the size of the features maps, reverting it back to the original input image dimensions. 
Further, DeepLab \cite{DeepLab} deployed dilated or atrous convolutions to achieve a multi-scale framework and increase the size of the receptive fields, avoiding downsampling in the network with the introduction of the Atrous Spacial Pyramid Pooling (ASPP). The ASPP architecture applies atrous convolutions in four parallel branches with different rates, and combines them via bilinear interpolation in order to recover the feature maps at the original image resolution.

Improving upon ASPP \cite{DeepLab}, the WASP module incorporates multi-scale features without immediately parallelizing the input stream \cite{WASPnet}, \cite{UniPose}. The WASP module creates a waterfall flow by initially processing through a filter and later creating a new branch. The waterfall scheme extends the cascade approach by combining the streams from all its branches to achieve a multi-scale representation. The OmniPose framework \cite{OmniPose} recently introduced the enhanced WASPv2 module, that improves upon the multi-scale  feature extraction from the backbone and includes the decoder features of the network.

\section{BAPose Architecture}\label{sec:bapose_architecture}
The proposed BAPose bottom-up method, illustrated in Figure \ref{fig:BAPose}, consists of a single-pass, single output branch network that is particularly effective for multi-person 2D pose estimation in crowded scenes.
BAPose integrates improvements in multi-scale  feature representations \cite{OmniPose}, \cite{DEKR}, an encoder-decoder structure combined with the spatial pyramid pooling of the waterfall configuration, and  disentangled adaptive regression for person localization and parts association.

The processing pipeline of the BAPose architecture is shown in Figure \ref{fig:BAPose}. The input image is initially processed by the HRNet feature extractor. The extracted multi-scale feature maps are then processed by the WASPv2 module with integrated decoder, that extracts the location of keypoints as well as contextual information for the localization regression. The network  generates $K$ heatmaps, one for each joint, with the corresponding confidence maps as well as 2 offset maps for the identification of person instances and association of keypoints to each instance.
The integrated WASPv2 decoder in our network generates detections from all scales of the feature extraction for both visible and occluded joints while maintaining the image resolution through the network. 

\begin{figure*}
\begin{center}
\includegraphics[width=1\linewidth]{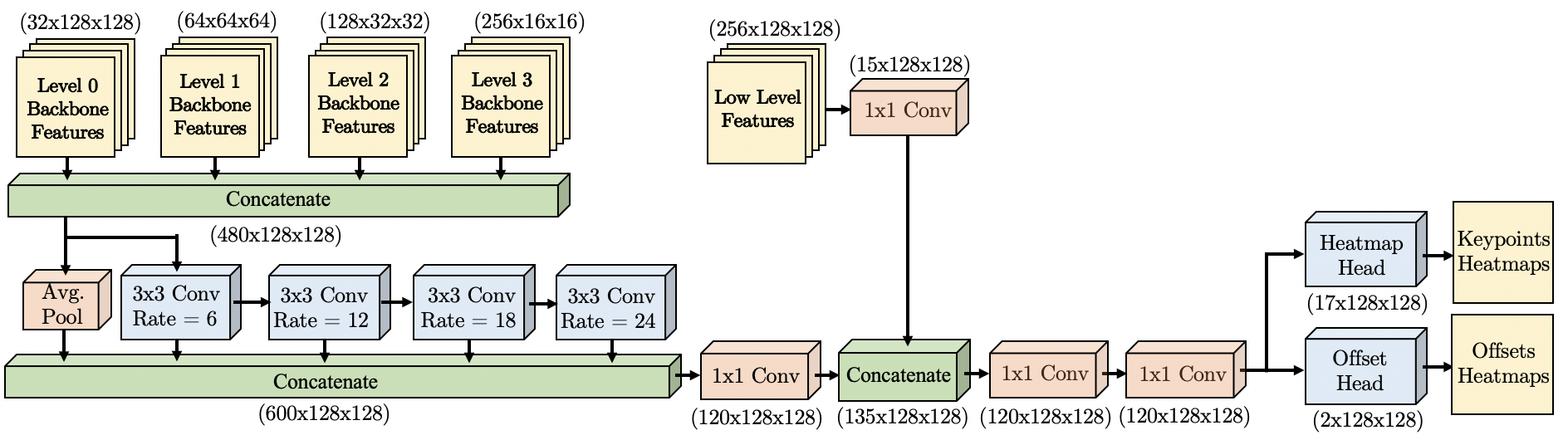}
\end{center}
  \caption{The proposed D-WASP disentangled waterfall module. The inputs are 32, 64, 128, and 256 features maps from all four levels of the HRNet backbone, respectively, and low-level features from the initial layers of the framework. The module outputs both the keypoints and offsets heatmaps.}
\label{fig:WASPv2}
\end{figure*}

Our architecture includes several innovations that contribute to increased accuracy. 
In the WASPv2 module, BAPose combines atrous convolutions and the waterfall architecture to increase the network's capacity to represent multi-scale contextual information by the probing of feature maps at multiple rates of dilation.
This configuration achieves a larger FOV in the encoder.
Our architecture also integrates disentangled adaptive convolutions in the decoding process, enabling the single-pass detection of multiple person instances and their keypoint estimation.
Additionally, our network demonstrates superior ability to deal with a large number of subjects by the enhanced extraction of features at multiple scales, as indicated by state-of-the-art results for the CrowdPose dataset presented in Section \ref{sec:results}.
Finally, the modular nature of BAPose facilitates the easy implementation and training of the network.

\subsection{Disentangled Waterfall Module}
The proposed enhanced ``Disentangled Waterfall Atrous Spatial Pyramid'' module, or D-WASP, is shown in Figure \ref{fig:WASPv2}.
The D-WASP module processes all four levels of feature maps from the backbone through the waterfall branches with different dilation rates. 
Low-level and high-level features are represented at the same resolution, achieving a refined localization for joint estimation.
Furthermore, the D-WASP module uses adaptive convolution blocks to infer the final heatmaps for joint localization and offset maps for person instance regression.
The module generates both the keypoints and offset heatmaps for each person, through their respective heads illustrated in Figure \ref{fig:WASPv2}. 
The D-WASP architecture helps to
more effectively discern multiple people in a crowded setting due to its multi-level and
multi-scale representations, contributing to SOTA performance.


The design of the D-WASP module relies on a combination of atrous and adaptive convolutions.
Atrous convolutions are utilized in the initial stages to expand the FOV by performing a filtering cascade at increasing rates to gain efficiency.
The waterfall modules are designed to create a waterfall flow, initially processing the input and then creating a new branch.
D-WASP goes beyond the cascade approach of \cite{DeepLabv3+} by combining all streams from all its branches and the average pooling layer from the original input.  
Additionally, our module incorporates a larger number of scales compared to WASPv2 \cite{OmniPose} by adopting all 480 feature maps from all levels of the HRNet feature extractor.
Adaptive convolutions are used to better infer the individual keypoints and offset heatmaps during the regression process by providing context around the vicinity of each detected joint and strengthening the relationship between associated joints.

\subsubsection{Waterfall Features and Adaptive Convolutions}
The D-WASP module operation begins with the concatenation $g_{0}$ of all feature maps $f_{i}$ from the HRNet feature extractor, where $i=0,1,2,3$ indicates the levels at different scales of the feature extractor and summation is overloaded for concatenation:
\vspace{-0.1in}
\begin{equation}
    g_{0} = \sum_{i=0}^{3}(f_{i})
\end{equation}

Following the concatenation of all feature maps, the waterfall processing is described as follows:
\begin{equation}
    f_{Waterfall} = W_1\circledast(\sum_{i=1}^{4}(W_{d_i}\circledast g_{i-1})+AP(g_{0}))
\end{equation}
\begin{equation}   
    f_{maps} = W_1\circledast(W_1\circledast(W_1\circledast f_{LLF} + f_{Waterfall})
\end{equation}
\noindent 
where $\circledast$ represents convolution, $g_{0}$ is the input feature map, $g_{i}$ is the feature map resulting from the $i^{th}$ atrous convolution, $AP$ is the average pooling operation, $f_{LLF}$ are the low-level feature maps, and $W_1$ and $W_{d_i}$ represent convolutions of kernel size 1$\times$1 and 3$\times$3 with dilations of ${d_i}=[1,6,12,18]$, as shown in Figure \ref{fig:WASPv2}. 
After concatenation, the feature maps are combined with low level features.
The last 1$\times$1 convolution brings the number of feature maps down to one quarter of the number in the combined input feature maps.

Finally, the D-WASP module output $f_{D-WASP}$ is obtained from the multi-scale adaptive convolutional regression, where adaptive convolution is defined as:
\begin{equation}   
    {\bf y}(c) = \sum_{i=1}^{9}({\bf w_i} {\bf x} (g_i^c+c))
\end{equation}
\noindent 
where $c$ is the center pixel of the convolution, ${\bf y}(c)$ represents the output of the convolution for input {\bf x}, ${\bf w_i}$ are the kernel weights for the the center pixel its neighbors, and $g_i^c$ is the offset of the $i^{th}$ activated pixel.
In the adaptive convolutions, the offsets $g_i^c$
are adopted in a parametric manner as an extension of spatial transformer networks \cite{STN}.

\subsubsection{Disentangled Adaptive Regression}
The regression stage for multi-person pose estimation is considered the most challenging and a bottleneck in performance for bottom-up methods.
To address the limitation of regression, additional processing may utilize pose candidates, post-processing matching schemes, proximity matching, and statistical methods, however these may be computationally expensive or limited in effectiveness.

D-WASP expands on the idea of regression by focus, by not only learning disentangled representations for each of the $K$ joints, but also using multiple scales to infer each representation for all keypoints from multiple adaptively activated pixels. This configuration gives each regression a more robust contextual information of the keypoint region, and results in a more accurate spatial representation.

The multi-scale approach proposed by the D-WASP module, allows BAPose to regress person detections and keypoints with a larger FOV, increasing the network capability to infer joints association through the use of adaptive convolutions. Differently than the WASPv2 \cite{OmniPose} decoder stage that only extracts the heatmaps for joints, the D-WASP multi-scale disentangled adaptive regression determines both the keypoint heatmaps and the final offset heatmaps that are used to regress the position of each individual in the image and their respective joints.

In addition, the integration of the multi-scale feature maps in the disentangled adaptive regression utilizes multiple resolutions at the regression stage, allowing the network to better infer the locations of people and their joints in the image. As a consequence, BAPose demonstrates superior performance (see Section \ref{sec:results}), especially in challenging scenarios that include large numbers of people in close proximity.

\section{Datasets}\label{sec:datasets}
We evaluated the BAPose method on two datasets for 2D multi-person pose estimation: Common Objects in Context (COCO) \cite{COCO} and CrowdPose \cite{CrowdPose}.  
The large and most commonly adopted COCO dataset \cite{COCO} consists of over 200K images with more than 250K instances of labelled people keypoints. The keypoint labels consist of 17 keypoints including all major joints in the torso and limbs, as well as facial landmarks of nose, eyes, and ears. The dataset is considered a challenging dataset due to the large number of images in a diverse set of scales and occlusion for poses in the wild.

The CrowdPose dataset \cite{CrowdPose} is a more challenging dataset since it includes many images with crowds and low separation among individuals. The dataset contains 10K images for training, 2K images for validation, and 20K images for testing.
The dataset contains frames with joints annotations, head and torso orientations, and body part occlusions. We follow evaluation procedures adopted by \cite{HigherHRNet} and \cite{DEKR}.

We adopted the generation of ideal Gaussian maps for the joints ground truth locations in order to train our network more effectively. The Gaussian maps are a more effective strategy for loss assessment during training compared to single points at the joint locations. As a consequence, the BAPose was trained to generate heatmaps as output locations for each joint.
The value of $\sigma=3$ was adopted, generating a well define Gaussian response for both the ground truth and keypoint predictions, while maintaining a decent separation of keypoints and avoiding large overlapping of keypoints.

\section{Experiments}\label{sec:experiments}
BAPose experiments followed standard metrics set by each dataset, and same procedures applied by \cite{HigherHRNet}, and \cite{DEKR}.

\subsection{Metrics}
For the evaluation of BAPose, the evaluation is done based on the Object Keypoint Similarity metric (OKS).

\begin{equation}
    OKS = \frac{(\sum_i{e^{-d_i^2/2s^2k^2_i}})\delta(v_i>0)}{\sum_i\delta(v_i>0)}
\end{equation}

\noindent 
where, $d_i$ is the Euclidian distance between the estimated keypoint and its ground truth, $v_i$ indicates if the keypoint is visible, $s$ is the scale of the corresponding target, and $k_i$ is the falloff control constant.
Since the OKS measurement is adopted by both COCO and CrowdPose dataset and is similar to the intersection over the union (IOU), we report our OKS results as the Average Precision (AP) for the IOUs for all instances between 0.5 and 0.95 ($AP$), at 0.5 ($AP^{50}$) and 0.75 ($AP^{75}$), as well as instances of medium ($AP^M$) and large size ($AP^L$) for the COCO dataset. For the CrowdPose dataset, we report easy ($AP^E$), medium ($AP^M$,) and hard size ($AP^H$) instances. We also report the overall Average Recall (AR) as well as AR for $AR^M$ medium and $AR^L$ large instances.


\begin{figure*}[t]
\begin{center}
\includegraphics[width=0.95\linewidth]{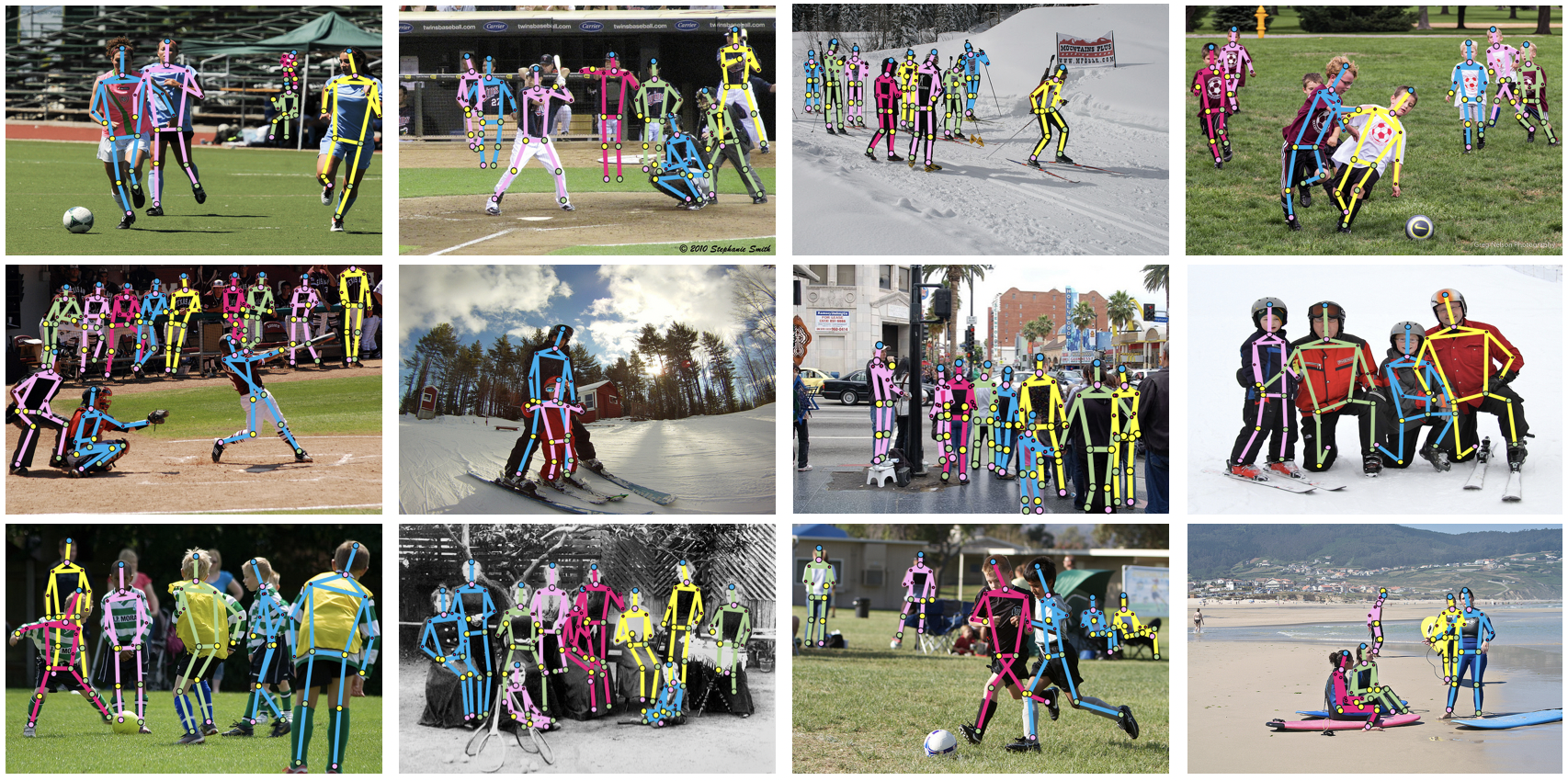}
\end{center}
  \caption{Pose estimation examples using BAPose with the CrowdPose dataset.}
\label{fig:CrowdPose_sample}
\end{figure*}

\begin{table*}[!ht]
\begin{center}
\begin{tabular}{|c|c|c|c|c|c|c|c|c|}
\hline
\multirow{2}{*}{Method}&
Input&
\multirow{2}{*}{Approach}&
\multirow{2}{*}{AP}&
\multirow{2}{*}{$AP^{50}$}&
\multirow{2}{*}{$AP^{75}$}&
\multirow{2}{*}{$AP^{E}$}&
\multirow{2}{*}{$AP^{M}$}&
\multirow{2}{*}{$AP^{H}$}\\
&Size&&&&&&&\\
\hline
\multicolumn{9}{|c|}{Single-Scale Testing}\\
\hline
\textbf{BAPose (W32)}&512&Bottom-Up&
\textbf{72.2\%}&\textbf{89.6\%}&\textbf{78.0\%}&\textbf{79.9\%}&\textbf{73.4\%}&\textbf{61.3\%}\\
HRNet-W48 \cite{DEKR}&640&Bottom-Up&
67.3\%&86.4\%&72.2\%&74.6\%&68.1\%&58.7\%\\
HigherHRNet-W48 \cite{HigherHRNet}&640&Bottom-Up&
65.9\%&86.4\%&70.6\%&73.3\%&66.5\%&57.9\%\\
MIPNet \cite{MIPNet}&512&Top-Down&70.0\%&-&-&-&-&-\\
Joint-candidate SPPE  \cite{CrowdPose}&-&Top-Down&66.0\%&84.2&71.5&75.5\%&66.3\%&57.4\%\\
HRNet-W32 \cite{DEKR}&512&Bottom-Up&
65.7\%&85.7\%&70.4\%&73.0\%&66.4\%&57.5\%\\
AlphaPose \cite{AlphaPose}&-&
Bottom-Up&-&-&-&71.2\%&61.4\%&51.1\%\\
Mask R-CNN \cite{MaskRCNN}&-&
Bottom-Up&60.3\%&-&-&69.4\%&57.9\%&45.8\%\\
OpenPose \cite{OpenPose}&-&Bottom-Up&-&-&-&62.7\%&48.7\%&32.3\%\\
\hline
\end{tabular}
\end{center}
\caption{BAPose results and comparison with SOTA methods for the CrowdPose dataset for testing.}
\label{tab:CrowdPoseval}
\end{table*}

\subsection{Parameter Selection}
We use a set of dilation rates of $r =$ \{1, 6, 12, 18\} for the D-WASP module, similar to \cite{OmniPose}.
The network was trained for 140 epochs. The learning rate is calculated based on the step method, where the rate is initialized at $10^{-3}$ and is reduced by an order of magnitude in two steps at 90 and 120 epochs. The training procedure includes random rotation between $-30^{\circ}$ and $30^{\circ}$, random scale from $0.75$ to $1.5$, and random translation between $-40$ and $40$, mirroring procedures followed by \cite{DEKR}. 
All experiments were performed using PyTorch on Ubuntu 16.04. The workstation has an Intel i5-2650 2.20GHz CPU with 16GB of RAM and an NVIDIA Tesla V100 GPU.

\section{Results}\label{sec:results}
This section presents BAPose results on two large datasets and provides comparisons with state-of-the art methods.

\subsection{Experimental results on the CrowdPose dataset}
We performed training and testing on the CrowdPose dataset, which presents a difficult challenge due to the high occurrence of crowds in the images. The CrowdPose results are shown in Table \ref{tab:CrowdPoseval}.

Our BAPose method significantly improves upon the performance of SOTA methods for 512$\times$512 input resolution,
achieving 72.2\% accuracy, and  significantly outperforms other bottom-up approaches, even those that utilized higher input resolutions. It is noticeable that BAPose achieves most of its gains by more precise joint estimations, increasing the performance from 70.4\% to 78.0\% for $AP^{75}$ when compared to the previous SOTA, HRNet-w32 \cite{DEKR}.
Additionally, BAPose outperforms networks that utilize top-down approaches. Differently than top-down methods, BAPose does not rely on ground truth for person detection and has to infer the location of all individuals in a modular, single-pass process. For the CrowdPose dataset, BAPose's performance is superior to networks utilizing higher resolution inputs of 640$\times$640 \cite{DEKR}, \cite{HigherHRNet} 
while processing the less computationally expensive 512$\times$512 resolution.

\begin{figure*}[ht!]
\begin{center}
\includegraphics[width=1\linewidth]{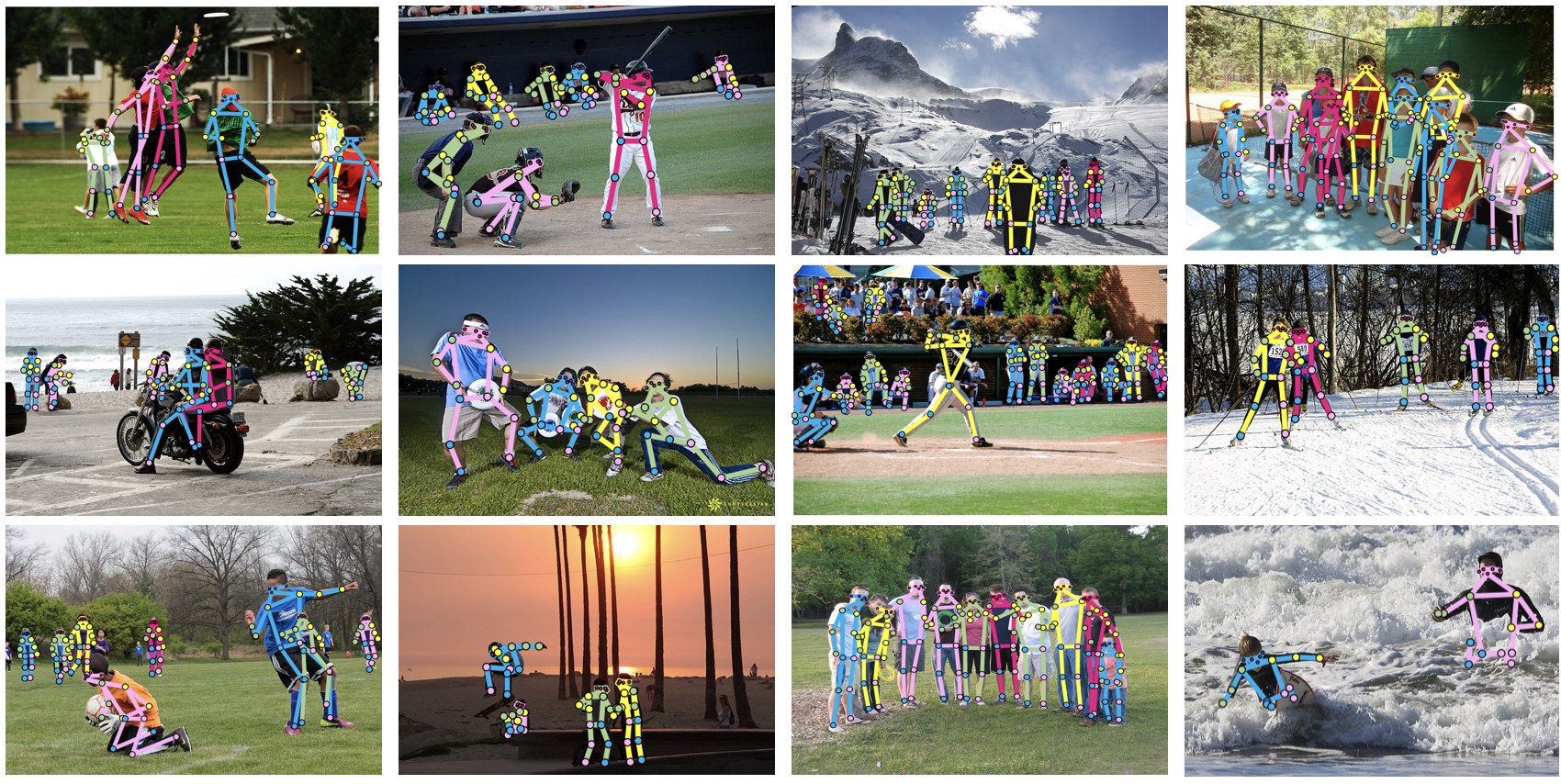}
\end{center}
  \caption{Pose estimation results using BAPose with the COCO dataset.}
\label{fig:COCO_sample}
\end{figure*}

Figure \ref{fig:CrowdPose_sample} illustrates successful detections of pose for multiple people in images from the CrowdPose test set.
The examples demonstrate how effectively BAPose deals  with occlusions and close proximity of individuals, as well as detections at different scales.

\subsection{Experimental results on the COCO dataset}
We next performed training and testing on the COCO dataset, which is challenging due to the large number of diverse images with multiple people in close proximity, and additionally includes images lacking a person instance.

We first compared BAPose with SOTA methods for the COCO validation and test-dev datasets, with results presented in Tables \ref{tab:COCOval} and \ref{tab:COCOtest-dev} respectively. The validation results in Table \ref{tab:COCOval} show that BAPose achieves significant improvement over the previous SOTA for input resolution of 512$\times$512 and 640$\times$640.
The BAPose results at the lower 512$\times$512 resolution are obtained with a significantly lower computational cost compared to methods with higher resolution inputs, as shown in Table \ref{tab:GFLOPs}, while achieving comparable results to the higher resolution. 

\begin{table}
\begin{center}
\begin{tabular}{|c|c|c|c|}
\hline
\multirow{2}{*}{Method}&Input&\multirow{2}{*}{GFLOPs}&Params\\
&Size&&(M)\\
\hline
HRNet-W32 \cite{DEKR}&512&45.4&29.6\\
\textbf{BAPose-W32}&512&56.8&30.3\\
HRNet-W48 \cite{DEKR}&640&141.5&65.7\\
HigherHRNet-W48 \cite{HigherHRNet}&640&154.3&63.8\\
\textbf{BAPose-W48}&640&183.2&67.4\\
AE \cite{AE}&512&206.9&227.8\\
PersonLab \cite{PersonLab}&1401&405.5&68.7\\
\hline
\end{tabular}
\end{center}
\caption{GFLOPs and number of parameters comparison.}
\label{tab:GFLOPs}
\end{table}

The combination of the HRNet backbone with the D-WASP module achieves an increased overall accuracy of 69.1\% when using single-scale testing, and 71.9\% when using multi-scale testing, compared to the previous SOTA, HRNet-w32 \cite{DEKR}, of 68\% and 70.7\%, respectively. Overall BAPose achieves a significant increase of 1.6\% and 1.7\% for single-scale and multi-scale testing, respectively.

BAPose improves the accuracy of the previous SOTA in all keypoint estimation metrics and IOU for the COCO dataset. Most of the performance improvements of BAPose are attributed to performing better on harder detections and more refined predictions at $AP^{75}$. The  results on the COCO validation dataset, in Table \ref{tab:COCOval}, show the greater capability of BAPose to detect more complex and harder poses while still using a smaller resolution in the input image.
We also trained and tested BAPose-W48 at 640$\times$640 resolution. BAPose-W48 achieved 71.5\% accuracy for the COCO validation set with single scale testing and 72.7\% with multi-scale testing, improving the previous SOTA by 0.7\% and 0.6\%, respectively.
However, larger resolution models require much higher computational resources, as illustrated by the GFLOPs and memory requirements for different methods shown in Table \ref{tab:GFLOPs}.
Compared to BAPose-W32, HRNet-W48 requires 249.1\% the number of GFLOPs and HigherHRNet-W48 requires 271.7\% the number of  GFLOPs, demonstrating that BAPose-W32 results in a better trade-off between accuracy and computational cost.

\begin{table*}[!ht]
\begin{center}
\begin{tabular}{|c|c|c|c|c|c|c|c|c|c|}
\hline
\multirow{2}{*}{Method}&Input&\multirow{2}{*}{AP}&\multirow{2}{*}{$AP^{50}$}&\multirow{2}{*}{$AP^{75}$}&\multirow{2}{*}{$AP^{M}$}&\multirow{2}{*}{$AP^{L}$}&\multirow{2}{*}{$AR$}&\multirow{2}{*}{$AR^M$}&\multirow{2}{*}{$AR^L$}\\
&Size&&&&&&&&\\
\hline
\multicolumn{10}{|c|}{Single-Scale Testing}\\
\hline
\textbf{BAPose (W48)}&640&
\textbf{71.5\%}&\textbf{88.7\%}&\textbf{77.8\%}&\textbf{67.2\%}&\textbf{78.8\%}&\textbf{76.2\%}&\textbf{71.0\%}&\textbf{84.0\%}\\
HRNet-W48 \cite{DEKR}&640&
71.0\%&88.3\%&77.4\%&66.7\%&78.5\%&76.0\%&70.6\%&84.0\%\\
HigherHRNet-W48 \cite{HigherHRNet}&640&
69.9\%&87.2\%&76.1\%&-&-&-&65.4\%&76.4\%\\
PersonLab \cite{PersonLab}&1401&66.5\%&86.2\%&71.9\%&62.3\%&73.2\%&70.7\%&65.6\%&77.9\%\\
PersonLab \cite{PersonLab}&601&
54.1\%&76.4\%&57.7\%&40.6\%&73.3\%&57.7\%&43.5\%&77.4\%\\
\hline
\textbf{BAPose (W32)}&512&
\textbf{69.1\%}&\textbf{87.0\%}&\textbf{75.6\%}&\textbf{63.1\%}&\textbf{78.6\%}&\textbf{73.7\%}&\textbf{66.9\%}&\textbf{83.4\%}\\
HRNet-W32 \cite{DEKR}&512&
68.0\%&86.7\%&74.5\%&62.1\%&77.7\%&73.0\%&66.2\%&82.7\%\\
HigherHRNet-W32 \cite{HigherHRNet}&512&
67.1\%&86.2\%&73.0\%&-&-&-&61.5\%&76.1\%\\
HGG \cite{HGG}&512&
60.4\%&83.0\%&66.2\%&-&-&64.8\%&-&-\\
CenterNet-HG \cite{CenterNet}&512&
64.0\%&-&-&-&-&-&-&-\\
CenterNet-DLA \cite{CenterNet}&512&
58.9\%&-&-&-&-&-&-&-\\
\hline
\multicolumn{10}{|c|}{Multi-Scale Testing}\\
\hline
\textbf{BAPose (W48)}&640&
\textbf{72.7\%}&\textbf{88.6\%}&\textbf{79.1\%}&\textbf{69.3\%}&78.4\%&\textbf{77.9\%}&\textbf{73.4\%}&84.7\%\\
HRNet-W48 \cite{DEKR}&640&
72.3\%&88.3\%&78.6\%&68.6\%&\textbf{78.6\%}&77.7\%&72.8\%&\textbf{84.9\%}\\
HigherHRNet-W48 \cite{HigherHRNet}&640&
72.1\%&88.4\%&78.2\%&-&-&-&67.8\%&78.3\%\\
\hline
\textbf{BAPose (W32)}&512&
\textbf{71.9\%}&\textbf{88.3\%}&\textbf{77.8\%}&\textbf{67.2\%}&\textbf{79.1\%}&\textbf{76.6\%}&\textbf{71.3\%}&\textbf{84.5\%}\\
HRNet-W32 \cite{DEKR}&512&
70.7\%&87.7\%&77.1\%&66.2\%&77.8\%&75.9\%&70.5\%&83.6\%\\
HigherHRNet-W32 \cite{HigherHRNet}&512&
69.9\%&87.1\%&76.0\%&-&-&-&65.3\%&77.0\%\\
HGG \cite{HGG}&512&
68.3\%&86.7\%&75.8\%&-&-&72.0\%&-&-\\
\hline
\end{tabular}
\end{center}
\caption{BAPose results and comparison with SOTA methods for the COCO dataset for validation.}
\label{tab:COCOval}
\end{table*}

\begin{table*}[!ht]
\begin{center}
\begin{tabular}{|c|c|c|c|c|c|c|c|c|c|}
\hline
\multirow{2}{*}{Method}&Input&\multirow{2}{*}{AP}&\multirow{2}{*}{$AP^{50}$}&\multirow{2}{*}{$AP^{75}$}&\multirow{2}{*}{$AP^{M}$}&\multirow{2}{*}{$AP^{L}$}&\multirow{2}{*}{$AR$}&\multirow{2}{*}{$AR^M$}&\multirow{2}{*}{$AR^L$}\\
&Size&&&&&&&&\\
\hline
\multicolumn{10}{|c|}{Single-Scale Testing}\\
\hline
\textbf{BAPose (W48)}&640&
\textbf{70.3\%}&\textbf{89.6\%}&\textbf{77.5\%}&\textbf{65.9\%}&\textbf{77.1\%}&\textbf{75.4\%}&\textbf{69.8\%}&\textbf{83.2\%}\\
HRNet-W48 \cite{DEKR}&640&
70.0\%&89.4\%&77.3\%&65.7\%&76.9\%&75.4\%&69.7\%&83.2\%\\
HigherHRNet-W48 \cite{HigherHRNet}&640&
68.4\%&88.2\%&75.1\%&64.4&74.2&-&-&-\\
PersonLab \cite{PersonLab}&1401&66.5\%&88.9\%&72.6\%&62.4\%&72.3\%&71.0\%&66.1\%&77.7\%\\
\hline
\textbf{BAPose (W32)}&512&
\textbf{68.0\%}&\textbf{88.0\%}&\textbf{74.8\%}&\textbf{62.4\%}&\textbf{76.6\%}&\textbf{72.9\%}&\textbf{66.1\%}&\textbf{82.4\%}\\
HRNet-W32 \cite{DEKR}&512&
67.3\%&87.9\%&74.1\%&61.5\%&76.1\%&72.4\%&65.4\%&81.9\%\\
SPM \cite{SSPM}&-&
66.9\%&88.5\%&72.9\%&62.6\%&73.1\%&-&-&-\\
PifPaf \cite{Pifpaf}&-&
66.7\%&-&-&62.4\%&72.9\%&-&-&-\\
$MDN_3$ \cite{MDR}&-&
62.9\%&85.1\%&69.4\%&58.8\%&71.4\%&-&-&-\\
CenterNet-HG \cite{CenterNet}&512&
63.0\%&86.8\%&69.6\%&58.9\%&70.4\%&-&-&-\\
OpenPose \cite{OpenPose}&-&
61.8\%&84.9\%&67.5\%&57.1\%&68.2\%&66.5\%&-&-\\
CenterNet-DLA \cite{CenterNet}&512&
57.9\%&84.7\%&63.1\%&52.5\%&67.4\%&-&-&-\\
AE \cite{AE}&512&
56.6\%&81.8\%&61.8\%&49.8\%&67.0\%&-&-&-\\
\hline
\multicolumn{10}{|c|}{Multi-Scale Testing}\\
\hline
\textbf{BAPose (W48)}&640&
\textbf{71.2\%}&\textbf{89.4\%}&\textbf{78.1\%}&\textbf{67.4\%}&76.8\%&\textbf{76.8\%}&\textbf{71.6\%}&\textbf{84.0\%}\\
HRNet-W48 \cite{DEKR}&640&
71.0\%&89.2\%&78.0\%&67.1\%&\textbf{76.9\%}&76.7\%&71.5\%&83.9\%\\
HigherHRNet-W48 \cite{HigherHRNet}&640&
70.5\%&89.3\%&77.2\%&66.6\%&75.8\%&-&-&-\\
Point-set Anchors \cite{Point-setAnchors}&640&
68.7\%&89.9\%&76.3\%&64.8\%&75.3\%&74.8\%&69.6\%&82.1\%\\
PersonLab \cite{PersonLab}&1401&
68.7\%&89.0\%&75.4\%&64.1\%&75.5\%&75.4\%&69.7\%&83.0\%\\
\hline
\textbf{BAPose (W32)}&512&
\textbf{70.4\%}&\textbf{89.3\%}&\textbf{77.4\%}&\textbf{66.0\%}&\textbf{76.9\%}&\textbf{75.6\%}&\textbf{70.1\%}&\textbf{83.2\%}\\
HRNet-W32 \cite{DEKR}&512&
69.6\%&89.0\%&76.6\%&65.2\%&76.5\%&75.1\%&69.5\%&82.8\%\\
HGG \cite{HGG}&512&
67.6\%&85.1\%&73.7\%&62.7\%&74.6\%&71.3\%&-&-\\
AE \cite{AE}&512&
65.5\%&86.8\%&72.3\%&60.6\%&72.6\%&70.2\%&64.6\%&78.1\%\\
\hline
\end{tabular}
\end{center}
\caption{BAPose results and comparison with SOTA methods for the COCO dataset for test-dev.}
\label{tab:COCOtest-dev}
\end{table*}

Figure \ref{fig:COCO_sample} presents examples of pose estimation results for the COCO dataset. It is noticeable from the sample images that BAPose effectively locates symmetric body joints and avoids confusion due to occlusion between individuals. This is illustrated in harder to detect joints such as ankles and wrists.
Overall, the BAPose results demonstrate its robustness for pose estimation in various challenging conditions, such as images that include detections of individuals with high overlapping ratio combined with shadows or darker images, or partial pose present in the image.

We next compared BAPose with SOTA methods on the larger COCO test-dev dataset, with results shown in Table \ref{tab:COCOtest-dev}. BAPose again achieved  new SOTA performance over methods using input resolutions of 512$\times$512. Our method obtained an overall precision of 68.0\% when using single-scale testing and 70.4\% when using multi-scale testing, which are improvements of 1.0\% and 1.1\% over SOTA for single and multi scale testing, respectively.
When training and testing at the 640$\times$640 resolution, BAPose-W48 achieved accuracies of 70.3\% for single-scale testing and 71.2\% when using single-scale multi-scale testing, an improvement of 0.4\% and 0.3\% to the previous SOTA, respectively. 
These results further confirm  that BAPose demonstrates most significant improvements in smaller and harder targets consistent with the findings from the validation dataset.

\section{Conclusion}\label{sec:conclusion}
We presented the BAPose method for bottom-up multi-person pose estimation. The BAPose network includes the novel D-WASP module that combines multi-scale features obtained from the waterfall flow with the person detection capability of the  disentangled adaptive regression. 
BAPose is a end-to-end trainable, single-pass architecture that does not require anchor poses, prior person detections, or postprocessing.
The results demonstrate state-of-the-art performance for both the COCO and CrowdPose datasets using various metrics, and the superior capability of person detection and pose estimation in densely populated images.

\section{Acknowledgements}
This research was supported in part by the National Science Foundation grant $\sharp$1749376.

{\small
\bibliographystyle{ieee_fullname}
\bibliography{egbib}
}

\end{document}